%% file: main.tex
\documentclass[11pt,a4paper,dvipsnames]{article}
\usepackage[dvipsnames]{xcolor}
\usepackage[hyperref]{acl2022}
\usepackage{amsmath}
\usepackage{times}
\usepackage{latexsym}
\usepackage{linguex}
\usepackage{xspace}
\usepackage{graphicx}
\usepackage{inconsolata}
\usepackage{tikz-dependency}
\usepackage{cleveref}
\usepackage{booktabs}
\usepackage{multirow}
\usepackage{algorithm}
\usepackage{algpseudocode}
\usepackage{amssymb}
\usepackage{makecell}
\usepackage{subcaption}
\usepackage{pythonhighlight}

\usepackage{microtype}

\input{defs.tex}

\title{Weakly Supervised Headline Dependency Parsing}

\author{Adrian Benton\thanks{\ \ Now at Google Research. Work done while at Bloomberg and Cornell University, respectively.} \\
  Bloomberg \\ \And
  Tianze Shi$^*$  
  \\
  Cornell University \\
  \texttt{\{adbenton,tianze\}@google.com} \quad\quad\quad\quad\quad\quad\quad\quad\quad\quad \\ \And
  Ozan İrsoy \\
  Bloomberg \\ 
  \quad\quad\quad\quad\quad\quad\quad\quad\quad\quad\quad \texttt{\{oirsoy,imalioutov\}@bloomberg.net} \\ \And
  Igor Malioutov \\
  Bloomberg \\
}

\date{}

\begin{document}

\maketitle

\begin{abstract}

English news headlines form a register with unique syntactic properties that have been documented in linguistics literature since the 1930s.  However, headlines have received surprisingly little attention from the NLP syntactic parsing community. We aim to bridge this gap by providing the first news headline corpus of Universal Dependencies annotated syntactic dependency trees, which enables us to evaluate existing state-of-the-art dependency parsers on news headlines. To improve English news headline parsing accuracies, we develop a projection method to bootstrap silver training data from unlabeled news headline-article lead sentence pairs. Models trained on silver headline parses demonstrate significant improvements in performance over models trained solely on gold-annotated long-form texts. Ultimately, we find that, although projected silver training data improves parser performance across different news outlets, the improvement is moderated by constructions idiosyncratic to outlet.

\end{abstract}

\input{introduction}

\input{background}

\input{data}

\input{silver}

\input{training_regime}

\input{intrinsic_results}

\input{extrinsic_results}

\input{related_work}

\input{conclusion}

\input{limitations}

\input{ack}

\bibliographystyle{acl_natbib}
\bibliography{main}

\clearpage
\appendix
\input{appendix.tex}

\end{document}

%% file: defs.tex
\newcommand{\removed}[1]{}
\newcommand{\REMOVED}[1]{}

\newcommand{\num}{\texttt{NUM}}

\definecolor{PredicateColor}{RGB}{255,102,102}

\newcommand{\predicate}[1]{$[_{P \,}  $\textbf{\underline{\color{PredicateColor} #1}}$]$}
\newcommand{\argone}[1]{$[_{A1 \,}  $\emph{\color{Blue} #1}$]$}
\newcommand{\argtwo}[1]{$[_{A2 \,}  $\emph{\color{YellowOrange} #1}$]$}
\newcommand{\argthree}[1]{$[_{A2 \,}  $\emph{\color{Green} #1}$]$}

\newcommand{\rel}[1]{\texttt{#1}\xspace}

\newcommand{\algvar}[1]{\textsf{#1}\xspace}

\newcommand{\TheData}{EHT\xspace}

\newcommand{\result}[2]{\ensuremath{#1_{\pm #2}}}
\newcommand{\resultbest}[2]{\ensuremath{\mathbf{#1}_{\pm #2}}}

%% file: introduction.tex
\section{Introduction}
\label{sec:introduction}

English news headlines are written to convey the most salient piece of information in an article in as little space as possible. This makes them an attractive target for information extraction systems, and other NLP applications that operate on the most salient information in a news article. Headlines have been the target for many NLP tasks including semantic clustering \cite{wities2017consolidated,laban2021news}, multi-document summarization \cite{bambrick-etal-2020-nstm}, and sentiment/stance classification \cite{strapparava2007semeval,ferreira2016emergent}.

\begin{figure}
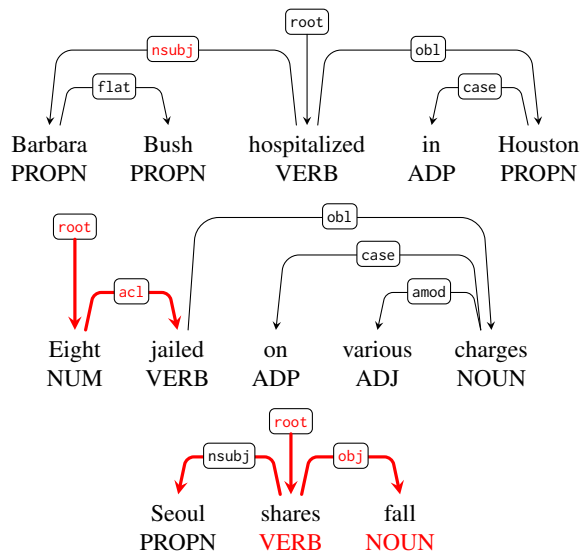

\centering
\begin{dependency}[theme = default,font=\footnotesize]
   \begin{deptext}[column sep=1em]
      Barbara \& Bush \& hospitalized \& in \& Houston \\ 
      PROPN \& PROPN \& VERB \& ADP \& PROPN \\
   \end{deptext}
   \depedge[edge above]{3}{1}{{\color{red}\rel{nsubj}}}
   \depedge[edge above]{1}{2}{\rel{flat}}
   \deproot[edge above,edge height=1.6cm]{3}{\rel{root}}
   \depedge[edge above]{5}{4}{\rel{case}}
   \depedge[edge above]{3}{5}{\rel{obl}}
\end{dependency}
\begin{dependency}[theme = default,font=\footnotesize]
   \begin{deptext}[column sep=1em]
      Eight \& jailed \& on \& various \& charges \\
      NUM \& VERB \& ADP \& ADJ \& NOUN \\
   \end{deptext}
   \deproot[edge above,edge style={red!70!red,very thick},edge height=1.6cm]{1}{{\color{red}\rel{root}}}
   \depedge[edge above,edge style={red!70!red,very thick}]{1}{2}{{\color{red}\rel{acl}}}
   \depedge[edge above]{5}{3}{\rel{case}}
   \depedge[edge above]{5}{4}{\rel{amod}}
   \depedge[edge above]{2}{5}{\rel{obl}}
\end{dependency}
\begin{dependency}[theme = default,font=\footnotesize]
   \begin{deptext}[column sep=1em]
      Seoul \& shares \& fall \\
      PROPN \& {\color{red}VERB} \& {\color{red}NOUN} \\
   \end{deptext}
   \depedge[edge above,edge style={red!70!red,very thick}]{2}{1}{\rel{nsubj}}
   \deproot[edge above,edge style={red!70!red,very thick},edge height=1.2cm]{2}{{\color{red}\rel{root}}}
   \depedge[edge above,edge style={red!70!red,very thick}]{2}{3}{{\color{red}\rel{obj}}}
\end{dependency}
    \caption{Example English news headlines with parses and POS tags generated by Stanza \citep{qi2020stanza}. Mispredicted relations and labels in {\color{red}red}. The text shown is after truecasing, before being fed to Stanza.}
    \label{fig:headline_examples}
\end{figure}

However, while headlines often present the most salient information from an article, brevity introduces its own obstacles. English news headlines are written in a unique register known as \emph{headlinese}. The structure of this register is determined primarily by typographical constraints along with the various functions that headlines serve, including summarization and eliciting reader interest \cite{maardh1980headlinese}. Headlinese syntax deviates from long-form news body through such features as a preference for atypical word senses and terms, frequent omission of determiners and auxiliaries, the acceptability of nominal and adverbial phrases, as well as multiple independent phrases in a single headline (decks).  \Cref{fig:headline_examples} presents a sample of English headlines exhibiting some of these properties, and parse errors made by a strong English dependency parser, Stanza \cite{qi2020stanza}.

While the syntax of English headlines deviates significantly from article body text, there has been little work in evaluating and developing classical NLP pipeline models for this register. News headline-related NLP tasks, such as headline generation \cite{rush2015neural,takase2016neural,takase2019positional} and classification \cite{kozareva2007ua,oberlander2020goodnewseveryone}, do not rely on syntactic annotations like POS tags, syntactic or semantic parses. This is by design, as the oftentimes poor performance of existing syntactic parsers on headlinese has impeded their application to tasks such as sentence compression \cite{filippova2013overcoming}.

In this work, we take a step towards improving headline dependency parsers by releasing the first English news headline treebank annotated according to universal dependency (UD) typology. We present the first quantitative evaluation of existing dependency parsers on English headlinese, and we propose a method for generating weak supervision for headline dependency parsers inspired by cross-lingual annotation projection \cite{yarowsky-etal-2001-inducing}.

\paragraph{Contributions}


\begin{enumerate}
    \item We release the first English headline treebank of $1{,}055$ manually annotated and adjudicated universal dependency (UD) syntactic dependency trees, the \textbf{E}nglish \textbf{H}eadline \textbf{T}reebank (\TheData), to encourage research in improving NLP pipelines for English headlinese.\footnote{The \TheData, licensed under CC-4.0, is available at 
    \url{https://github.com/bloomberg/emnlp22\_eht}\xspace}
    \item We establish baselines on the \TheData evaluation set with existing state-of-the-art parsers. Our experiments confirm prior observations that existing syntactic parsers perform poorly on headlinese \cite{filippova2013overcoming}.
    \item We propose a tree projection method to generate weak supervision for training more accurate headline parsers, and demonstrate that training on silver-annotated trees can significantly reduce parsing errors. Most strikingly, we show that that after finetuning on weak supervision, we are able to reduce root prediction relative error rate by 92.8\% within domain, and by 21.3\% for an out-of-domain wire. We further show that these gains translate to downstream improvements in the quality of tuples extracted by an open domain information extraction system.
    
\end{enumerate}

This paper is structured as follows: \Cref{sec:background} presents prior work on linguistic analyses of headlinese and their treatment in the NLP community; \Cref{sec:data} describes the \TheData annotation process and descriptive statistics; \Cref{sec:projection} describes our tree projection algorithm for generating silver headline dependency trees; \Cref{sec:training_headline_parsers} describes our experiment set up; Sections 6 and 7 
respectively present intrinsic parser performance on the \TheData and extrinsic performance on an open information extraction (OpenIE) task; and \Cref{sec:related_work} presents related work on headline and low-resource syntactic processing.

%% file: background.tex
\section{Background}
\label{sec:background}

\paragraph{Linguistic Analysis of Headlines}
English news headlines are known for their compressed telegraphic style, constituting a unique register
known as \emph{headlinese} \cite{garst1933headlines,straumann1935newspaper}.
Through a manual corpus analysis of
over $1{,}800$ headlines from two British newspapers,
\citet{maardh1980headlinese} finds that
headlinese shares some syntactic features with ``ordinary'' English language,
but there also exist
a number of features peculiar to headlinese.
These include the validity of nominal and adverbial headlines,
lack of determiners, omission of auxiliaries and copulas,
and use of the present tense to denote urgency of the event.
Nevertheless, these syntactic hallmarks of headlinese vary across country of publication \cite{ehineni2014syntactic}, news outlet \cite{maardh1980headlinese,siegal1999new}, and time period \cite{vanderbergen1981grammar,schneider2000emergence,afful2014diachronic}, making development of a strong, general English headline parser particularly challenging.

\paragraph{Headline NLP}

In spite of the clear evidence that headlinese differs significantly from standard written English syntax,
there has been scant work on building traditional NLP pipeline components for headlines.
This has limited the linguistic features that NLP researchers can extract from headlines, and subsequently limited the analyses that can be performed on them.
For instance, \citet{filippova2013overcoming} reports that
poor parsing accuracy for headlines
impedes their use of headline parses in alignment with a body sentence.

This is not to say that headlines have been ignored as an object of study by the community.  Tasks such as headline generation, compression, and news summarization are all well-studied problems, partly because they circumvent the need for annotation of linguistic structure \cite{filippova2013overcoming,rush2015neural,tan2017neural,takase2019positional,ao2021pens}. Other studied tasks include emotion identification/sentiment analysis \cite{kozareva2007ua,oberlander2020goodnewseveryone}, stance identification \cite{ferreira2016emergent}, framing or bias detection \cite{gangula2019detecting,liu2019detecting}, and headline clustering \cite{laban2021news}.

%% file: data.tex
\section{The English Headline Treebank}
\label{sec:data}

Here we describe the compilation and characteristics of our evaluation set, the \textbf{E}nglish \textbf{H}eadline \textbf{T}reebank (\TheData).

\subsection{Data Sources and Pre-processing}
\label{subsec:data_sources}

\input{tables/basic-stats}

We sample English news headlines from two sources to build the \TheData: the Google sentence compression corpus \citep[GSC;][]{filippova2013overcoming}, and the New York Times Annotated Corpus\footnote{\url{https://catalog.ldc.upenn.edu/LDC2008T19}} (NYT).
We sample from the GSC as it contains hundreds of thousands of news headlines across tens of thousands of domains, and as is described in \Cref{sec:projection}, we leverage it as a rich source of silver-annotated training data. We sample $600$ headlines from GSC in total.

In addition, we sample from NYT to form an out-of-domain evaluation set, which was not subjected to the same preprocessing decisions used to build the GSC. We sample $500$ headlines uniformly at random from the NYT, under the constraint that they are 4 to 12 tokens long (up to the 95$^{\text{th}}$ percentile). We impose this length constraint to avoid trivial parses, as well as noise in the data.\footnote{For examples, occasionally the NYT headline field appeared to contain a full article body.} Of these $500$ headlines, we removed $45$ headlines that were templated death notices and obituaries.\footnote{e.g., "Paid Notice : Deaths GOLDBERG , HERBERT".}

All headlines are tokenized using the Stanford Penn Treebank tokenization algorithm with default settings. We use Stanza \cite{qi2020stanza} to bootstrap our expert annotators with predicted UD-style part-of-speech tags and parse trees.
To reduce the discrepancy between training data of Stanza and our news headline data, we truecased headlines using an n-gram truecaser model trained on English news body text, and inserted a period at the end of the headline before running inference with Stanza.
\Cref{tab:basic-stats} shows the number of headlines and tokens in our headline datasets.

\subsection{Annotation and Adjudication}
\label{subsec:annotation}

Our annotation follows the UD guidelines whenever possible. In \Cref{sec:appendix:annotation_guidelines},
we provide an addendum for consistent treatment of syntactic constructions
that frequently occur in headlines,
but are underspecified in the original UD guidelines.

In the first stage of annotation,
each headline is independently annotated for POS tag sequence and dependency parse by two expert annotators\footnote{
All four annotators and adjudicators are fluent English speakers, have a background in NLP,
and were trained in the UD guidelines before annotation.
}
using the UD Annotatrix interface \citep{tyers2017ud}.
Any discrepancies between annotators are resolved in the second, adjudication, stage.
Two adjudicators independently examine the annotations
and pick the one that conforms to UD guidelines,
or construct their own parse if they disagree with both candidate parses from the first stage.\footnote{
During adjudication, the identities of the first-stage annotators are anonymized to avoid biasing towards or against any particular annotator.}
The third and last stage is group discussion, where all four annotators and adjudicators discuss and resolve any remaining disagreements.

First-stage annotation takes roughly one minute per instance per annotator, and the combination of second and third-stage adjudication takes another minute per instance per adjudicator. On a sample of 50 headlines held out to compute inter-annotator agreement, we find that 56\% of headlines were parsed or POS-tagged incorrectly by Stanza. 48\% contain some attachment error and 50\% have an incorrect relation label. Annotators achieved a 72\% headline-level agreement rate on this sample in the first stage, with individual POS tag agreement of 98.8\% and labeled dependency attachment agreement rate of 94.8\%. Many annotator discrepancies arose from parsing the internal structure of named entities, which were resolved during the subsequent adjudication phases. These along with other common issues are listed in \Cref{sec:appendix:annotation_guidelines}.

\subsection{Characterizing Headline Data}
\label{subsec:characterizing_headline_data}

\begin{figure*}[t]
    \centering
    \includegraphics[width=\textwidth]{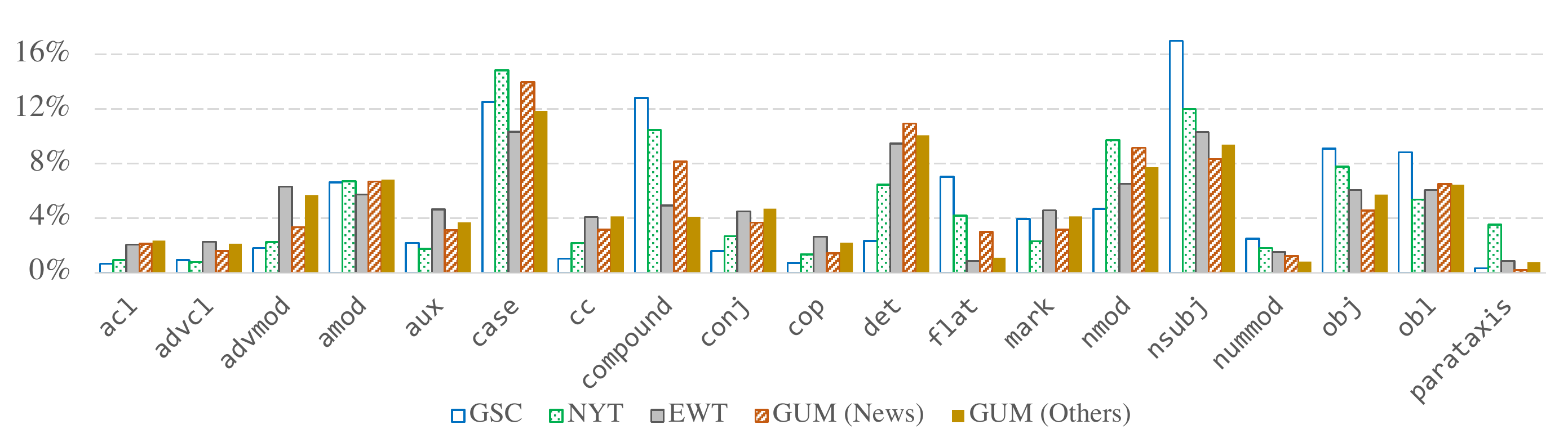}
    \caption{Distributions of dependency relation labels across the \TheData,
    compared with UD 2.8 EWT and GUM corpora.
    We exclude \rel{punct} and \rel{root} relations when calculating the distributions, and omit low-frequency labels (below $2\%$ across all datasets) in this chart.    
    }
    \label{fig:rel-stats}
\end{figure*}

\autoref{fig:rel-stats} presents the distributions of relation labels in the \TheData
compared to the UD 2.8 English Web Treebank \citep[EWT;][]{silveira2014gold}
and the Georgetown University Multilayer corpus \citep[GUM;][]{zeldes2017gum}.
The EWT includes data from
web media (weblogs, newsgroups, emails, reviews, and Yahoo! answers),
and the texts in GUM corpus
are drawn from a range of domains including news, fiction, academic writings, as well as dialogue such as transcribed interviews.

Compared with texts from other domains, 
English news headlines use fewer determiners, auxiliaries, and copulas,
which is consistent with prior linguistic characterization of headlinese \citep{maardh1980headlinese}.
News headlines have higher proportions of
\rel{compound} and \rel{flat} relations,
due to frequent mentions of named entities,
and
we also observe larger percentages of \rel{nsubj} and \rel{obj} relations,
as a consequence of headline brevity and focus on core argument structure.

%% file: tables/basic-stats.tex
\begin{table}[t]
    \centering
    \begin{tabular}{cl|rr}
    \toprule
     & Dataset & Headlines & Tokens \\
    \midrule
    \multirow{2}{*}{\TheData} & GSC & $600$ & $5{,}017$ \\
    & NYT & $455$ & $3{,}986$ \\
    Silver & Projection & $48{,}633$ & $395{,}237$ \\
    \bottomrule
    \end{tabular}
    \caption{Statistics of our gold (\TheData) data and silver (projected GSC) data.}
    \label{tab:basic-stats}
\end{table}

%% file: silver.tex
\section{Generating Silver Data by Projecting from Lead Sentences}
\label{sec:projection}

\begin{algorithm}[!t]
\caption{
Algorithm for projecting a parse tree from
a news article lead sentence \algvar{s} to a headline \algvar{h}, which is a subsequence of \algvar{s}.
}
\label{alg:projection1}
\begin{algorithmic}
\State \textbf{Definitions}
\State $\algvar{N(s)}$ : set of nodes in tree \algvar{s}, each corresponding to a word in the sentence or the dummy root
\State $\algvar{N(h)} \subseteq \algvar{N(s)}$ : set of nodes in tree \algvar{h}

\\
\Function{ExtractSubtree}{\algvar{s}, \algvar{h}}
  \State $\algvar{N'} \gets \varnothing$ \Comment{Subset of nodes to be returned}
  \ForAll{$\algvar{n} \in \algvar{N(h)}$}
    \State $\algvar{P}\gets$ nodes on path from root of \algvar{s} to \algvar{n}
    \State $\algvar{N'} \gets \algvar{N'} \cup \algvar{P}$
  \EndFor
  \State $\algvar{R'} \gets$ relations in tree \algvar{s} s.t. both the head and tail of the relation are in $\algvar{N'}$
  \State \Return $\algvar{N'}$, $\algvar{R'}$
\EndFunction

\\

\Function{Project}{\algvar{s}, \algvar{h}}
  \State $\algvar{N'}, \algvar{R'} \gets \textsc{ExtractSubtree}(\algvar{s}, \algvar{h})$
  \While{$|\algvar{N'}|>|\algvar{N(h)}|$}
    \State $\algvar{n}\gets$ closest to root s.t. $\algvar{n} \in \algvar{N'}, \algvar{n} \notin \algvar{N(h)}$
    \State $\algvar{c}\gets$ leftmost child of \algvar{n} s.t. $\algvar{c} \in \algvar{N'}$
    \State $\algvar{p}\gets$ parent of $\algvar{n}$ according to \algvar{R'}
    \State Update \algvar{R'}: attach all siblings of \algvar{c} to \algvar{c}
    \State and attach \algvar{c} to \algvar{p}
    \State $\algvar{N'} \gets \algvar{N'} \ \backslash \ \{\algvar{n}\}$
  \EndWhile
  \State \Return A tree formed by \algvar{N'}, \algvar{R'}
\EndFunction
\end{algorithmic}
\end{algorithm}

While \TheData is suitable for evaluating parser performance on English news headlines, $1{,}055$ headlines is much less data than is typically used for training a syntactic parser. For comparison, the EWT contains more than $15$ times as many tokens as the \TheData.

On the other hand, it may be data-inefficient to manually annotate a training set of tens of thousands of headlines, since English news headlines constitute a different register of written English, not a different language. Although certain constructions are idiosyncratic to headlines, one can often expand a headline to a well-formed sentence in the news body register, as words are frequently omitted to produce a headline \cite{straumann1935newspaper,maardh1980headlinese}.
This section describes an algorithm for automatically assigning dependency trees to unannotated headlines to create silver training data for training a headline dependency parser.

Our approach is based on the key observation that headlines convey similar semantic content as the bodies and they typically share many local sub-structures.
Lead sentences, often the first sentence in an article, serve a similar function as news headlines in grabbing reader attention and
stating essential facts about news events;
lead sentences are sometimes direct expansions of the headlines.
Consequently, the pairs of lead sentence and headline have been used to automatically construct examples for sentence compression \citep{filippova2013overcoming}.

\Cref{alg:projection1} projects the dependency tree annotation from a news article lead sentence to a headline, where the headline is a (possibly non-contiguous) subsequence of the lead sentence.
The main idea of this algorithm is to prune the lead sentence's dependency tree until it only contains those tokens in the headline.
When a token from the lead sentence is missing in the headline, but it has children appearing in both strings, we promote its first child to preserve connectivity.
For example, the following sentence snippet contains an extra ``promised'' than the corresponding headline:

\begin{dependency}[theme = default]
   \begin{deptext}
      Researchers \& \ldots \& promised \& to \& release \& data \& \ldots \\
   \end{deptext}
   \depedge[edge height=0.6cm]{3}{1}{\rel{nsubj}}
   \depedge[edge height=0.6cm]{3}{5}{\rel{xcomp}}
   \depedge[edge height=0.6cm]{5}{6}{\rel{obj}}
\end{dependency}

\noindent
and our algorithm promotes ``release'' to be the new root of the tree for the headline:

\begin{dependency}[theme = default]
   \begin{deptext}
      Researchers \& to \& release \& data \\
   \end{deptext}
   \depedge[edge height=0.6cm]{3}{1}{\rel{nsubj}}
   \depedge[edge height=0.6cm]{3}{4}{\rel{obj}}
\end{dependency}

We use \Cref{alg:projection1} to construct a silver-annotated corpus of headline dependency trees
from headline-lead sentence pairs in the GSC corpus.
Our silver corpus contains $48{,}633$ headlines that satisfied the subsequence constraint, the same magnitude as the EWT and significantly larger than
our manually-annotated \TheData. Of these, $8{,}633$ were held out as a development set, with the remaining $40{,}000$ used for training.

%% file: training_regime.tex
\section{Training Headline Parsers}
\label{sec:training_headline_parsers}

We vary two main dimensions during parser training: training data selection and data combination methods. We consider the EWT, the projected GSC headline data described in \Cref{sec:data}, and a combination of both as different training sets. We experiment with three different ways of combining training sets:
a) simply concatenate the two corpora;
b) use a multi-domain\footnote{
We abuse terminology here and use \emph{domain} to refer
to examples that come from different corpora, even if the distinction between language in each corpus is the register.
} model with shared feature extractors, but independent parameters for the parsing modules in each domain~\cite{benton2021cross};
and c) first train on the gold-standard EWT corpus,
and subsequently finetune on the silver-annotated GSC headline corpus.

\paragraph{Model}
Our model architecture follows the deep biaffine parser  \citep{dozat-manning17},
using a pre-trained BERT \citep{devlin2019bert} as a feature extractor. This architecture underlies many state-of-the-art dependency parsers \citep[e.g.,][]{kondratyuk-straka-2019-75} and the winning solutions in recent runs of IWPT shared tasks \citep{bouma-etal-2020-overview,bouma-etal-2021-raw}.
Model and implementation details are provided in \Cref{sec:appendix:hyperparameters}.

\paragraph{Combining EWT and Projected GSC}
In our experiments, we consider three different data combination methods:
\begin{enumerate}
    \item \textbf{Concat}: We simply concatenate the two corpora and train a dependency parser based on the joint dataset. This strategy does not require any modification to the model architecture or the training procedures.
    \item \textbf{MultiDom}: Inspired by the multi-domain POS tagging architecture in \citet{benton2021cross}, we experiment with a multi-domain parser. In this parsing architecture, we have one parser for EWT and another for headlines, sharing the same underlying BERT-based feature extractor. In other words, each parser has its own trainable projection and biaffine attention layers. In each training step, we sample a batch of examples from the concatenated corpora and jointly update the domain-specific parameters and shared feature extractor. 
    \item \textbf{Finetune}: Finally, we also experiment with a two-step training strategy where we finetune on the projected GSC headline data based on a trained parser on EWT. \citet{stymne-etal-2018-parser} find this strategy to be one of the most effective ways to learn from multiple treebanks in the same language.\footnote{They report another strategy of using treebank embeddings to be equally effective as finetuning, but that requires modification to the model by adding treebank embeddings and can be viewed as a simplification to our multi-domain parser.}
\end{enumerate}

\paragraph{Ensembling}

We train each parser under each setting with five random restarts and report means and standard deviations in \Cref{sec:intrinsic_results}.
To reduce variations in our manual analysis in \Cref{sec:extrinsic_results}, we analyze the ensembled parse trees using the reparsing technique of \citet{sagae-lavie-2006-parser}.

%% file: intrinsic_results.tex
\section{Intrinsic Parser Performance}
\label{sec:intrinsic_results}

\input{tables/table_main-results}

Intrinsic parser performance is shown in \Cref{tab:main_results_table}.  The discrepancy in baseline performance between NYT and GSC (85.49\% vs. 60.60\% LAS) can be attributed to the fact that NYT headlines exhibit a closer distribution of relation types to EWT than GSC headlines (\Cref{fig:rel-stats}).  Many NYT headlines already constitute a well-formed body sentence, albeit without final punctuation. This is further supported by the fact that only training on projected GSC parse trees significantly improves performance on GSC (89.09\% LAS) while actually hurting NYT performance, with a slight drop to 84.75\% LAS.

However, training on both EWT and projected GSC improves parser performance across both domains. We found that training a multi-domain model performed about as well as concatenating EWT and projected GSC training data. Ultimately, we found that a pipelined finetuning scheme -- first training on EWT, then silver projected GSC headlines -- yielded a strong parser across both domains (LAS of 87.13\% on NYT and 90.08\% on GSC).

\subsection{Error Analysis}
\label{subsec:error_analysis}

\begin{figure*}[t]
    \centering
    \includegraphics[width=\textwidth,page=4]{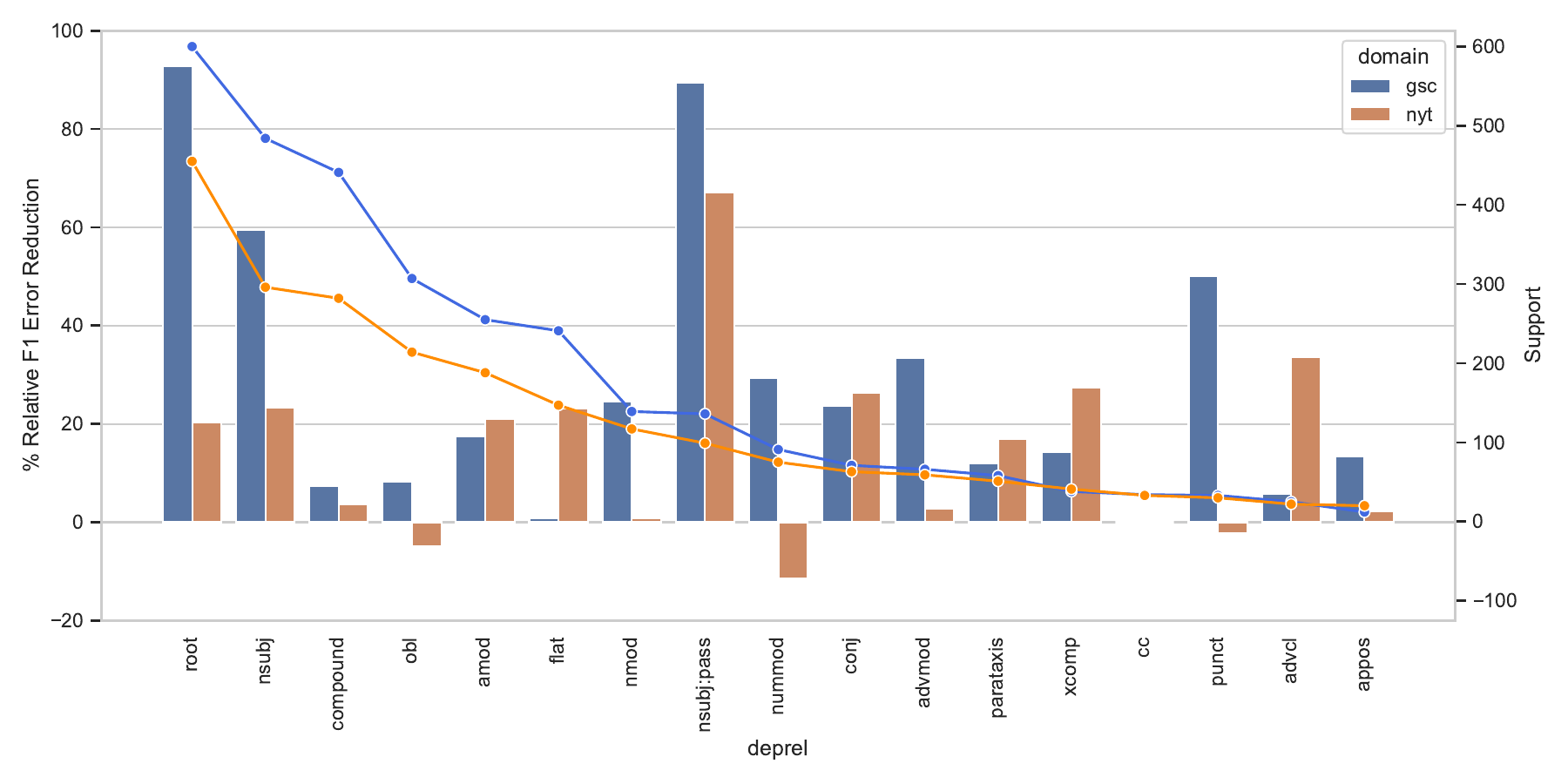} 
    \caption{\% relative error reduction in F1 score across dependency relations for both the GSC and NYT evaluation sets, from the ensembled \emph{Both (finetuning)} model to \emph{EWT (baseline)}. Relations are sorted by descending frequency and only relations that occurred at least 20 times in the evaluation set are shown. Support for each class is indicated by the line plot.
    }
    \label{fig:hdep_err_f1}
\end{figure*}

Although finetuning on GSC headlines with projected dependency parses improves parser performance on both the GSC and NYT evaluation sets, we see more marked improvements on the GSC corpus. \Cref{fig:hdep_err_f1} displays the \% relative error reduction in F1 of the GSC-finetuned ensemble against the EWT baseline ensemble broken by relation type. See \Cref{sec:appendix:f1_relation_errors} for absolute F1 for each relation type and domain. We compute model performance for all models using the \texttt{eval07.pl} evaluation script released as part of the First Shared Task on Parsing Morphologically-Rich Languages.\footnote{\url{http://www.spmrl.org/spmrl2014-sharedtask.html}}

\begin{figure*}
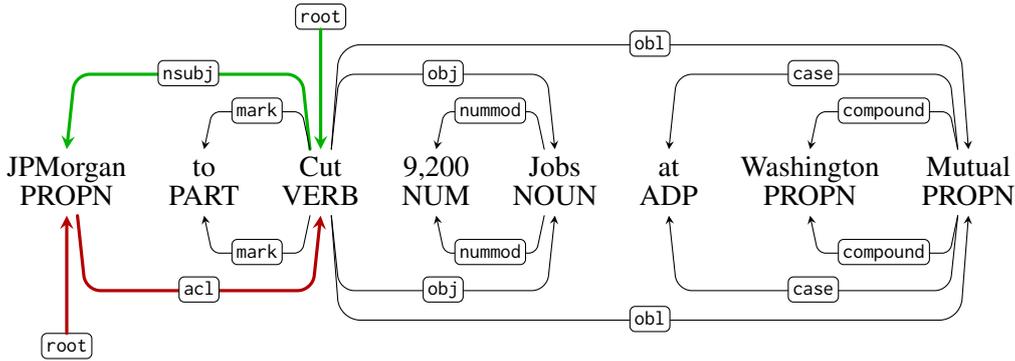

\centering
\begin{dependency}[theme = default]
   \begin{deptext}[column sep=1em]
      JPMorgan \& to \& Cut \& 9,200 \& Jobs \& at \& Washington \& Mutual \\
      PROPN \& PART \& VERB \& NUM \& NOUN \& ADP \& PROPN \& PROPN \\
   \end{deptext}
   \depedge[edge above,edge style={green!70!black,very thick}]{3}{1}{\rel{nsubj}}
   \depedge[edge above]{3}{2}{\rel{mark}}
   \deproot[edge above,edge style={green!70!black,very thick}]{3}{\rel{root}}
   \depedge[edge above]{5}{4}{\rel{nummod}}
   \depedge[edge above]{3}{5}{\rel{obj}}
   \depedge[edge above]{8}{6}{\rel{case}}
   \depedge[edge above]{8}{7}{\rel{compound}}
   \depedge[edge above,edge height=8ex]{3}{8}{\rel{obl}}
   \deproot[edge below,edge style={red!70!black,very thick}]{1}{\rel{root}}
   \depedge[edge below]{3}{2}{\rel{mark}}
   \depedge[edge below,edge style={red!70!black,very thick}]{1}{3}{\rel{acl}}
   \depedge[edge below]{5}{4}{\rel{nummod}}
   \depedge[edge below]{3}{5}{\rel{obj}}
   \depedge[edge below]{8}{6}{\rel{case}}
   \depedge[edge below]{8}{7}{\rel{compound}}
   \depedge[edge below,edge height=8ex]{3}{8}{\rel{obl}}
\end{dependency}
    \caption{Example parses given by \emph{EWT (baseline)} (bottom) and \emph{Both (finetuned)} (top) on an example headline from the GSC.  Differing edges are highlighted in green and red for finetuned and baseline, respectively.
    }
    \label{fig:anecdote_bad_root}
\end{figure*}

It is clear from the relation-level error analysis that most of the gains on GSC come from correct identification of the headline root, arguably the most important relation in the headline parse.  In fact, the finetuned parser achieves 98.2\% recall in identifying the root, whereas the baseline parser only achieves 74.6\% recall.  Headlines using the ``to VERB'' construction, indicating future tense or an expected event, are particularly susceptible to root misprediction by the baseline parser (example given in \Cref{fig:anecdote_bad_root}).  Performance on the \texttt{nsubj} relation also improves as a side effect of correctly identifying the root.

Gains on the NYT evaluation set are consistent across most relations, but with smaller improvements.  This is encouraging in that no NYT headline training data was used to train the model, silver or otherwise.  \texttt{parataxis} benefits from finetuning on projected GSC headlines. This relation occurs frequently in NYT headlines due to a preference for headlines with multiple decks, independent syntactic components: e.g., ``\emph{Essay ; B.C.C.I. : Justice Delayed}''.  The fact that this deck structure occurs more frequently in headlines results in a parser with a stronger prior for predicting \texttt{parataxis}.

It is also important to note that the finetuned parser can identify passive constructions much more accurately than the baseline. \% F1 performance for identifying \texttt{nsubj:pass} improves from 11.1\% to 90.5\%  on GSC and from 60.0\% to 86.8\% on NYT.

%% file: tables/table_main-results.tex
\begin{table*}
    \small
    \centering
        \begin{tabular}{l|c@{\hspace{3pt}}c@{\hspace{3pt}}c@{\hspace{3pt}}c|c@{\hspace{3pt}}c@{\hspace{3pt}}c@{\hspace{3pt}}c}
        \toprule
        \multirow{2}{*}{\makecell{\bf Training data / \\ \bf regime}} & \multicolumn{4}{c|}{\textbf{NYT}} & \multicolumn{4}{c}{\textbf{GSC}} \\ 
         & {\bf UAS} & {\bf LAS} & {\bf UEM} & {\bf LEM} & {\bf UAS} & {\bf LAS} & {\bf UEM} & {\bf LEM} \\ \midrule
        EWT & \result{88.82}{0.22} & \result{85.49}{0.28} & \result{57.89}{0.65} & \result{48.57}{1.03} & \result{83.01}{0.36} & \result{80.60}{0.28} & \result{50.80}{1.14} & \result{42.90}{0.68} \\
        Proj & \result{88.27}{0.30} & \result{84.75}{0.33} & \result{56.88}{1.17} & \result{48.22}{0.63} & \result{90.99}{0.23} & \result{89.09}{0.30} & \result{68.33}{0.59} & \result{59.93}{0.80} \\
        Concat & \resultbest{89.97}{0.65} & \result{87.05}{0.56} & \result{60.70}{1.35} & \result{53.14}{1.66} & \result{91.23}{0.19} & \result{89.32}{0.23} & \result{68.67}{0.67} & \result{61.23}{0.83} \\
        MultiDom & \result{89.58}{0.27} & \result{86.29}{0.36} & \result{60.00}{0.87} & \result{50.29}{0.90} & \result{91.16}{0.11} & \result{89.31}{0.21} & \result{68.63}{0.77} & \result{61.07}{1.00} \\
        Finetune & \result{89.93}{0.14} & \resultbest{87.13}{0.18} & \resultbest{61.45}{0.65} & \resultbest{53.71}{0.75} & \resultbest{91.88}{0.06} & \resultbest{90.08}{0.11} & \resultbest{71.07}{0.25} & \resultbest{63.37}{0.27} \\
        \bottomrule
    \end{tabular}
    \caption{
    Parsing accuracies on 
    NYT and GSC headlines from the \TheData, comparing models trained on
    EWT, silver headline projection data (Proj), and different methods
    for combining these two training data sources: concatenating (Concat), training with a multi-domain model (MultiDom), and finetuning on silver GSC headline trees (Finetune). UAS and LAS correspond to (un)labeled attachment score, and UEM/LEM to (un)labeled exact match score (at the sentence level). 
    }
    \label{tab:main_results_table}
\end{table*}

%% file: extrinsic_results.tex
\section{Extrinsic Evaluation}
\label{sec:extrinsic_results}

In addition to intrinsic evaluation of parsers, we also evaluate these models downstream. We perform an extrinsic evaluation using the state-of-the-art syntax-based PredPatt OpenIE System \cite{white2016universal}, and evaluate extracted tuples using the protocol and error typology taken from \citet{benton2021cross}. As PredPatt relies solely on a UD parse and POS tag sequence to extract candidate tuples, this constitutes a direct downstream evaluation of more accurate headline parses.

\paragraph{OpenIE Evaluation Protocol}
Two annotators independently annotated 200 extracted tuples manually.\footnote{A subset of the annotators from the dependency parse annotations.} These tuples were randomly sampled from the GSC and NYT headlines, such that the PredPatt extracted different OpenIE tuples from the baseline ensemble parse compared to finetuned ensemble parse. Each tuple was judged as either \emph{Correct}, or annotated with its most salient error type: \emph{Malformed Predicate}, \emph{Bad Sub-predicate}, \emph{Missing Core Argument}, \emph{Argument Misattachment}, or \emph{Incomplete Argument}.

To control for potential annotation bias, tuples were shuffled and identity of the parser and example domain were hidden from annotators. After independent annotation, the two annotators adjudicated conflicting annotations and converged on a single label for each tuple. Prior to adjudication, annotators achieved an agreement rate of 62\% for annotating salient error type, with a Cohen's $\kappa$ of 0.430. 
Many discrepancies in the first annotation round resulted from confusion between \emph{Malformed Predicate} and \emph{Argument Misattachment} or \emph{Bad Sub-predicate}. Often, several error types were present in the incorrect extractions, but deciding which error type was most salient was resolved during adjudication. Examples of each error type and annotation conventions are given in \Cref{sec:appendix:openie_typology}.

\input{tables/table_openie}

\paragraph{OpenIE Results} 
Results from a typological error analysis of 200 tuples are shown in \Cref{tab:openie_results}. As expected, \emph{Malformed Predicate}s were the predominant source of error for the baseline EWT model, followed by \emph{Missing Core Argument} errors. This agrees with the finding that headline root identification exhibited marked regressions in the baseline.

In our experiments, the domain-specific model was able to drastically reduce errors for both of these error types.
We registered a statistically significant improvement ($26\%$ absolute) in valid tuple extraction performance when using the output of the model finetuned on EWT+GSC data when compared to the EWT-only baseline. For NYT, the improvement was not statistically significant. We hypothesize that this is due to the fact that the wire exhibits more structural similarities to long-form text, as evidenced by the frequency of relation types (\Cref{fig:rel-stats}).

%% file: tables/table_openie.tex
\begin{table*}
\small
\begin{tabular}{ll|p{1.6cm}p{1.8cm}p{1.8cm}p{2.2cm}p{1.8cm}r}
\toprule
&    &  \textbf{Malformed Predicate} &  \textbf{Bad Sub-predicate} &  \textbf{Missing Core Argument} &  \textbf{Argument Misattachment} &  \textbf{Incomplete Argument} &  \textbf{Correct} \\
\textbf{Domain} & \textbf{Model} &                 &              &               &                &                 &                    \\
\midrule
GSC & EWT &            20 &         \textbf{4} &          14 &           4 &            2 &               56 \\
& Finetune  &            \textbf{4$^{\dagger}$} &         6 &          \textbf{2$^{*}$} &           6 &            \textbf{0} &               \textbf{82$^{\dagger}$} \\
NYT & EWT &            \textbf{10} &         8 &          6 &           12 &            \textbf{0} &               64 \\
& Finetune &            12 &         \textbf{6} &          \textbf{2} &           \textbf{4} &            2 &               \textbf{74} \\
\bottomrule
\end{tabular}
\caption{
\% error type for OpenIE tuples.  Statistically significantly better performance within domain, according to a two population proportion test is indicated by $*$ at the $p=0.05$ level and $\dagger$ at the $p=0.01$ level. Sample size of 50 tuples for each (domain, model) pair. Best performing model per (error type, domain) in bold.
}
\label{tab:openie_results}
\end{table*}

%% file: related_work.tex
\section{Further Related Work}
\label{sec:related_work}

\paragraph{Headline Syntactic Processing}

Perhaps the two most relevant works are the recently published \emph{POSH} \cite{benton2021cross} and \emph{GoodNewsEveryone} corpora \cite{oberlander2020goodnewseveryone}. \emph{POSH} is a dataset of POS-tagged English news headlines, without gold dependency parse annotations. \emph{GoodNewsEveryone}, on the other hand, contains thousands of emotion-bearing headlines labeled for semantic roles (SRL).  In \emph{GoodNewsEveryone}, the relationships between identified actor, target, and predicate are solely determined by their roles. Collecting dependency parse annotations is much more involved than either POS tagging or SRL, as dependency parses require identifying deep relationships between individual words that are not solely derived from their types.  That said, the release of both of these corpora underscores the importance of headlines as an object of study in NLP, and the desire for richer linguistic annotations.

\paragraph{Low Resource Syntactic Processing}
Low resource syntactic parsing is typically motivated by the need to develop a parser for languages with scant gold supervision \cite{vania2019systematic}. \citet{agic2016multilingual} employ a similar, yet more involved method of annotation projection to project parser predictions from a high-resource language to a low-resource language.  As we are not projecting across languages, and we restrict our parallel text to cases where a headline is a subsequence of the lead sentence, we rely on heuristics to repair the projected dependency parse.

Dependency parsers and treebanks for tweets are similar in spirit to the current work \cite{owoputi2013improved,kong2014dependency,liu2018parsing}. Unlike Tweebank, we chose not to develop our own annotation scheme, but rather annotate under the UD schema.
UD is sufficiently expressive for annotating headlines, and allows us to leverage multiple domains for training a parser.

%% file: conclusion.tex
\section{Conclusion}
\label{sec:conclusion}

In this work, we describe the first gold-annotated evaluation set for English headline UD dependency parsing, the \TheData. We hope this data will encourage further research in improving dependency parsers for overlooked registers of English. In addition, we hope that the development of accurate headline dependency parsers will result in stronger performance at existing headline understanding and processing tasks, and enable more subtle linguistic analysis, such as identification of ``crash blossom'' news headlines.

%% file: limitations.tex
\section*{Limitations}
\label{sec:limitations}

\paragraph{Variation across news outlets} 
Figure 3 demonstrates out-of-domain generalization to NYT headlines on several structurally important relations such as \texttt{root} and \texttt{nsubj:pass} by training on silver projected trees. However, some relations that can be more accurately predicted in GSC headlines do not generalize to NYT. These include adjunct relations such as \texttt{nmod}, \texttt{nummod}, and \texttt{advmod}. Even within a register as niche as English headlinese, there is significant variation in convention between news outlets.

\paragraph{General news headline distributions} The GSC corpus is originally collected by \citet{filippova2013overcoming} and contains crawled news headlines and lead sentences from a wide variety of news outlets. The headline-lead-sentence pairs are filtered to include only grammatical and informative headlines (see Section 4 of \citet{filippova2013overcoming}) and thus the resulting GSC corpus may not be representative of all English news headlines. The NYT corpus contains samples from a single outlet and is also not representative of the general news headline distribution.

\paragraph{Different news headline categories} Depending on the types of news articles (\textit{e.g.}, front page, editorials, op-eds, \textit{etc.}), their corresponding headlines may exhibit distinctive structural properties. Our work is agnostic to the different categories of news articles and their headlines.

\paragraph{Multilinguality} In this work, we demonstrate that training on silver parse trees projected onto English news headlines results in more accurate English headline parsers. For other languages, headlines may or may not exhibit significant grammatical differences from English headlines, and when they do, the types of headline constructions are language- and culture-dependent. We expect the benefits of training on projected trees to be mediated by the discrepancy between the ``conventional'' and headline grammar within a given language. In addition, for languages with richer morphology, morphological analysis may be required to align dependency relation annotations from a body sentence to its headline. As we only consider English headlines in this work, further exploration is required before determining whether the projection algorithm,
Algorithm 1, can be adapted to morphologically rich languages.

%% file: ack.tex

\section*{Acknowledgements}

We thank the anonymous reviewers for their insightful reviews.
Tianze Shi acknowledges support from Bloomberg's Data Science Ph.D. Fellowship.

%% file: appendix.tex
\section{Syntactic Annotation Guidelines}
\label{sec:appendix:annotation_guidelines}

\subsection*{General principles}

Please refer to the UD annotation guidelines (\url{https://universaldependencies.org/guidelines.html}) for general rules of syntactic dependency annotation. This document serves as an addendum to the UD guidelines, in order to detail how to annotate certain frequently occurring and/or headline-specific constructions.

\subsection*{Frequent headline constructions}

\paragraph{Headlines with multiple decks/components}

Use parataxis to connect multiple components. For example:

\ex.  Paid Notice: Deaths BROOKS, JOHN N.\\

includes three independent components: "Paid Notice", "Deaths", and "BROOKS, JOHN N.", with the latter two attached to the first through parataxis. There can be nested parataxis if necessary to reflect hierarchical structures within the headline components.

\paragraph{Headlines with omitted auxiliaries}

For frequent constructions including "NP VP$_\text{ed}$", "NP VP$_\text{ing}$", and "NP VP$_\text{to}$", where the finite auxiliary "be" verbs are omitted, we still treat the headlines as verbal headlines and mark the main verbs as the root/head of the headlines.

\paragraph{Reported speech}

Refer to the UD guidelines. Typically, a \rel{ccomp} or \rel{parataxis} relation is used.

\paragraph{\rel{flat} and \rel{compound}}

First, refer to UD guidelines on \rel{flat} and \rel{compound}. These are typically annotated as flat:

\begin{itemize}\itemsep0em 
    \item (Person) Names
    \item Company/team/organization/... names without internal (compositional) structures. (e.g., "Rolling Stones" is not compositional and should be analyzed as flat.)
    \item Foreign phrases
    \item Dates without explicit internal structures (excluding "the 1st of May")
    \item Titles/honorifics
\end{itemize}

\paragraph{Dates}

\ex.  Thursday, December 7, 2000\\

Refer to English web treebank example \texttt{email-enronsent06\_01-0005}

\vspace{1em}

\begin{dependency}[theme = default]
   \begin{deptext}[column sep=1em]
      Thursday \& , \& December \& 7 \& , \& 2000 \\
   \end{deptext}
   \depedge{1}{3}{\rel{appos}}
   \depedge{3}{4}{\rel{nummod}}
   \depedge{1}{6}{\rel{nmod:tmod}}
\end{dependency}

\paragraph{Currency}

\ex.  \$ 200 million\\

Refer to English web treebank example: \texttt{newsgroup-groups.google.com\_FOOLED\_ 1bf9cdc5a4c2ac48\_ENG\_ 20050904\_130400-0022}

\vspace{1em}

\begin{dependency}[theme = default]
   \begin{deptext}[column sep=1em]
      \$ \& 200 \& million \\
   \end{deptext}
   \depedge{1}{3}{\rel{nummod}}
   \depedge{3}{2}{\rel{compound}}
\end{dependency}

\paragraph{Game Scores}

\ex.  4 - 0 \\

Refer to English web treebank example: \texttt{newsgroup-groups.google.com\_hiddennook\_ 1fd8f731ae7ffaa0\_ENG\_ 20050214\_192900-0006}

\vspace{1em}

\begin{dependency}[theme = default]
   \begin{deptext}[column sep=1em]
      4 \& - \& 0 \\
   \end{deptext}
   \depedge{1}{3}{nmod}
   \depedge{3}{2}{case}
\end{dependency}

\subsection*{Special cases}

\paragraph{Named Entities}

Locations should be annotated similarly to people names, with \rel{flat}. Therefore "Lake Erie" should be parsed using \rel{flat} rather than \rel{compound}. Although this conflicts with how locations are annotated in EWT, the judgments in EWT are occasionally inconsistent or conflict with the UD annotation guidelines, as evidenced by UD issue 777 -- \url{https://github.com/UniversalDependencies/docs/issues/777}. \rel{flat} should also be used for names of racing horses, where although there is often compositional structure in these names, they are treated as a single unit as there are loose syntactic constraints on what constitutes a valid race horse name.

Company names with typical suffixes like "Inc." or "Co." should be analyzed with that word as the head, with a \rel{compound} relation to the idiosyncratic part of the company name. Arbitrary names of companies should be analyzed with \rel{flat}.
Names of creative works: visual art, books, movies, video games should be annotated such that internal structure is preserved, e.g., "Lord of the Rings" is not parsed with \rel{flat}.

\paragraph{Legitimate PP Attachment Ambiguity}

In certain cases there may be inherent ambiguity in where a prepositional phrase should attach, but the syntactic ambiguity has little effect on the meaning of the headline (\rel{nmod} on an oblique/object argument vs. \rel{obl} attaching to the matrix verb). In these cases, we chose to attach the PP as an \rel{nmod} to the argument, out of convention.

\paragraph{Hyphenated Words}

In the case of hyphenated words, we analyze the internal structure of the hyphenated words and attach as the entire hyphenated word functions in the headline. For example, "shake-up" is parsed as:

\vspace{1em}

\begin{dependency}[theme = default]
   \begin{deptext}[column sep=1em]
      shake \& - \& up \\
   \end{deptext}
   \depedge{1}{3}{\rel{compound:prt}}
\end{dependency}

even if the entire word functions as a noun.

\paragraph{Typos}

In the case of typographical errors or issues with data processing, we assume the intended word during annotation. So, for "Baby game changer for to", "to" is labeled as "\num", assuming "two" was the intended word. These typographical errors will be remedied by their corrected lemma in the future.

\section{Implementation of the Projection Algorithm}

\begin{figure*}
    \centering
\begin{python}
def project(heads: List[int], rels: List[str], subset: List[int]):
  """Finds the subtree spanned by the nodes in `subset`.
  
  Arguments:
    heads: Each element `heads[i]` corresponds to the head of node `i`;
      node 0 is root, and `heads[0] = -1`.
    rels: Each element `rels[i]` is the dependency relation label between
      `heads[i]` and node `i`.
    subset: A list of nodes that are found in the subtree.

  Returns:
    A tuple of heads and relations in the same format as `heads` and `rels`,
    the length of each is equal to `len(subset)`.
  """
  heads = deepcopy(heads)
  rels = deepcopy(rels)
  
  # Collect all involved nodes (ExtractSubtree).
  included = set()
  for i in subset:
    cur = i
    while cur != -1:
      included.add(cur)
      cur = heads[cur]

  # Cache the children of each node.
  children = [[] for x in heads]
  for i in sorted(included):
    if heads[i] != -1:
      children[heads[i]].append(i)

  while len(included) != len(subset):
    # Find the top-most node that is not currently in the subset.
    queue = [-1]
    while len(queue):
      cur = queue.pop()
      if cur not in subset and cur != -1:
        node_to_collapse = cur
        break
      queue.extend(children[cur])

    # Find the local structure and collapse.
    children_nodes = children[node_to_collapse]
    leftmost = children_nodes[0]
    for c in children_nodes:
      heads[c] = leftmost
    heads[leftmost] = heads[node_to_collapse]
    rels[leftmost] = rels[node_to_collapse]
    included.discard(node_to_collapse)

    # Update cache.
    children = [[] for x in heads]
    for i in sorted(included):
      if heads[i] != -1:
        children[heads[i]].append(i)

  # Extract the subgraph.
  mapping = {n: i for i, n in enumerate(subset)}
  subset_heads = [mapping.get(heads[x] , -1) for x in subset]
  subset_rels = [rels[x] for x in subset]

  return subset_heads, subset_rels
\end{python}
    \caption{The python implementation of \Cref{alg:projection1}.}
    \label{fig:projection-code}
\end{figure*}

\Cref{fig:projection-code} provides a detailed python implementation of \Cref{alg:projection1} for reproducibility.

\section{Parser and Implementation Details}
\label{sec:appendix:hyperparameters}

Our parser architecture combines the deep biaffine parser \citep{dozat-manning17} with the pre-trained contextual BERT feature extractor \citep{devlin2019bert}.
For words with multiple subword tokens, we adopt BERT representations on the final subword tokens.
For the deep biaffine parser, the attachment and labeling probabilities are determined by biaffine attention scores between pairs of head-dependent words,
which in turn are linearly projected from BERT embeddings and then followed by a non-linear leaky ReLU activation function.
We used a dimension of $400$ for the attachment biaffine scorer, and $100$ for the label scorer.
For the BERT feature extractor, the weights are initialized from the public bert-base-uncased model,\footnote{\url{https://huggingface.co/bert-base-uncased}} consisting of roughly 110 million parameters, and fine-tuned during training. Each model was trained on a single Nvidia GTX 2080 Ti GPU, and took up to two hours to train depending on when training was halted.

We selected learning rate on the baseline EWT model,\footnote{From the set of \{$5\times10^{-6}$, $10^{-5}$, $2\times10^{-5}$, $5\times10^{-5}$\}.} and used the same hyperparameter settings when training all other parsers.
We used a maximum learning rate of $10^{-5}$, a batch size of $8$, and a learning rate schedule that tenths the base learning rate every $5$ iterations without increase in validation accuracy, up to two times maximum. Learning rate was warmed up according to a linear schedule during the first $320$ iterations. Gradients are clipped to a maximum norm of $5.0$. We used Adam optimizer \citep{kingma-ba15} with $\beta_1=0.9$, $\beta_2=0.999$ for all training runs. For finetuning, we used a maximum learning rate of $10^{-6}$, with an identical learning rate schedule. Dropout rates of $0.3$ are applied to all non-linear activations in the parsing modules.

\section{Intrinsic Performance by Relation}
\label{sec:appendix:f1_relation_errors}

\begin{figure*}[t]
    \centering
    \begin{subfigure}{0.95\textwidth}
      \includegraphics[width=\textwidth]{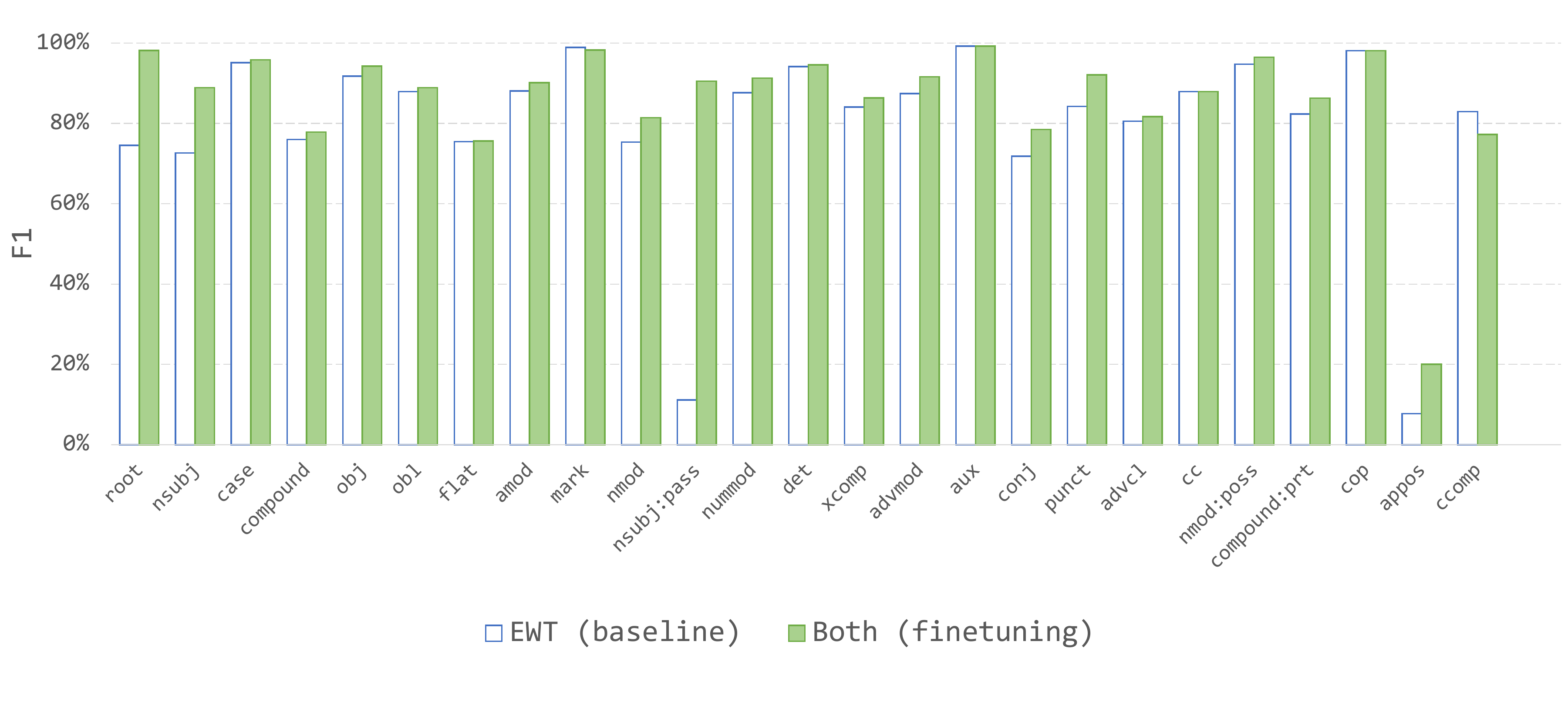}
    \end{subfigure}
    \begin{subfigure}{0.95\textwidth}
      \includegraphics[width=\textwidth]{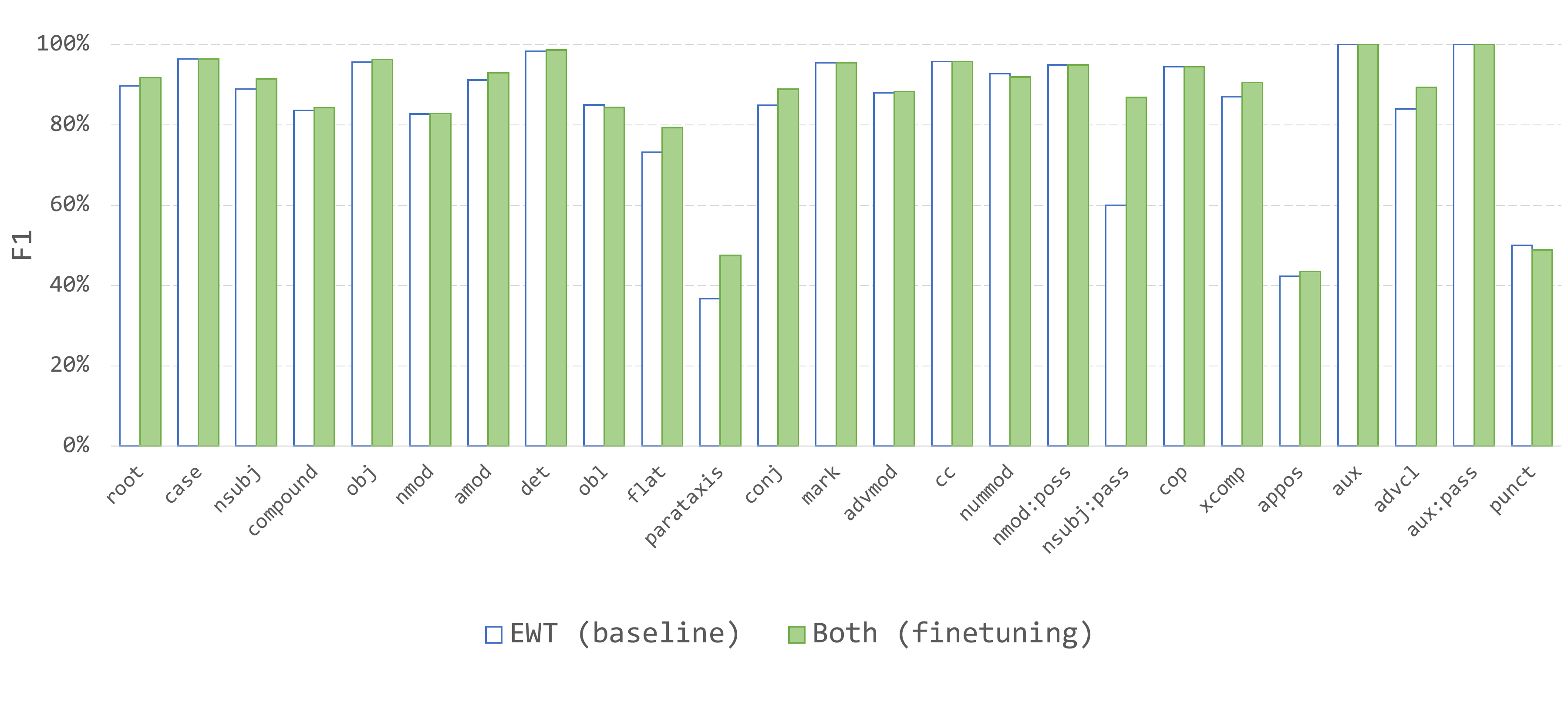}
    \end{subfigure}
    \caption{\% F1 score across dependency relations for both the GSC (top) and NYT (bottom) evaluation sets for the ensembled EWT-only model against the finetuned EWT+projected GSC predictions. Relations are sorted by descending frequency and only relations that occurred at least 20 times in the evaluation set are shown. 
    }
    \label{fig:hdep_absolute_f1}
\end{figure*}

Per-relation absolute F1 is displayed \Cref{fig:hdep_absolute_f1} for the ensembled baseline EWT-trained parser vs. additionally finetuning on projected GSC trees.

\section{OpenIE Annotation Details}
\label{sec:appendix:openie_typology}

\begin{table*}
    \centering
    \Large
\begin{center}
\begin{tabular}{l} 
 \toprule
 \textbf{Malformed Predicate} \\
 \midrule
  \argone{Torrid heatwave sweeps} \predicate{Punjab} \\
  \predicate{Kenya acrobat falls during} \argone{circus show in Moscow} \\
  \predicate{Will Wright to leave} \argone{Electronic Arts} \\ \midrule
 \textbf{Missing Core Argument} \\
 \midrule
  Toyota to \predicate{revise} \argone{dollar forecast to 80 yen} \\
  Several paths available to \predicate{extend} \argone{the litigation} \\ \midrule
 \textbf{Bad Sub-Predicate} \\
 \midrule
  \argone{Sanofi} to\predicate{take} \argtwo{control of Shantha Biotechnics} \\ \midrule
 \textbf{Argument Misattachment} \\ \midrule
  \argone{J.} \argtwo{W. Kirby} \predicate{wed to} \argthree{miss McCabe} \\
  \argone{Bishop} \argtwo{who} \predicate{had denied} \argthree{Holocaust} apologizes \\ \midrule
 \textbf{Incomplete Argument} \\ \midrule
  A raft of \argone{plans} \argtwo{that} \predicate{try to dispel} \argthree{math anxieties} \\
\bottomrule
\end{tabular}
\end{center}
\caption{Example salient error types for OpenIE tuples extracted from the baseline EWT-only ensemble parse.  Extracted tuples encoded as $\langle$\predicate{Predicate}, \argone{Argument 1}, \argtwo{Argument 2}$\rangle$.}
  \label{tab:openie_error_examples}
\end{table*}

\Cref{tab:openie_error_examples} contains a handful of examples for each of the salient OpenIE error types annotated in \Cref{sec:extrinsic_results}. Please refer to \citet{benton2021cross} for descriptions of each of these error types. In addition to that protocol, we adopted the following annotation conventions in order to consistently annotate corner cases. In general, tuples were labeled as incorrect \emph{only} if there were clear mistakes in the definition of arguments or predicate:

\begin{itemize}
  \item Tuples where the substructure of the reported phrase is decomposed as additional arguments in reporting structures (``Prime Minister says...'') are judged ``Correct''.
  \item A complicated predicate was labeled as valid, even when an object could have been treated as a separate argument.
  \item Tuples of independent decks, related by parataxis, or appositives are judged ``Correct''.
  \item Sub-predicates that are entailed by the headline are judge ``Correct''. For example, ``X engaged to wed'' $\rightarrow$ (wed, X) ; (engaged, X) ; (engaged to wed, X) are all valid.
  \item Relative pronouns should not be included as separate arguments in the relative clause, as they are redundant with the nominal head.
\end{itemize}